\documentclass[conference]{IEEEtran}
\IEEEoverridecommandlockouts
\usepackage{cite}
\usepackage{amsmath,amssymb,amsfonts}
\usepackage{algorithmic}
\usepackage{graphicx}
\usepackage{textcomp}
\usepackage{xcolor}
\usepackage{fontawesome}
\def\BibTeX{{\rm B\kern-.05em{\sc i\kern-.025em b}\kern-.08em
    T\kern-.1667em\lower.7ex\hbox{E}\kern-.125emX}}
\begin{document}

\title{Sentence Punctuation for Collaborative Commentary Generation in Esports Live-Streaming}

\author{\IEEEauthorblockN{Hong Huang\IEEEauthorrefmark{1}, Junjie H. Xu\IEEEauthorrefmark{1}}
\IEEEauthorblockA{\textit{Graduate School of} \\\textit{Comprehensive Human Sciences} \\
\textit{University of Tsukuba}\\
Tsukuba, Ibaraki, Japan \\
\{s2165071, s2021705\}@s.tsukuba.ac.jp}
\and
\IEEEauthorblockN{Xiaoling Ling}
\IEEEauthorblockA{\textit{Faculty of Arts} \\
\textit{University of British Columbia}\\
Vancouver, British Columbia, Canada \\
xling0@student.ubc.ca}
\and
\IEEEauthorblockN{Pujana Paliyawan}
\IEEEauthorblockA{\textit{Research Organization of} \\\textit{Science and Technology} \\
\textit{Ritsumeikan University}\\
Kusatsu, Shiga, Japan \\
pujana.p@gmail.com}
}

\maketitle
\begingroup\renewcommand\thefootnote{\IEEEauthorrefmark{1}}
\footnotetext{Equal contribution}
\endgroup
\begin{abstract}
To solve the existing sentence punctuation problem for collaborative commentary generation in Esports live-streaming, this paper presents two strategies for sentence punctuation for text sequences of game commentary, that is, punctuating sentences by two or three text sequence(s) originally punctuated by \textit{Youtube} to obtain a complete sentence of commentary.  
We conducted comparative experiments utilizing and fine-tuning a state-of-the-art pre-trained generative language model among two strategies and the baseline to generate collaborative commentary.
Both objective evaluations by automatic metrics and subjective analyses showed that our strategy of punctuating sentences by two text sequences outperformed the baseline.
\end{abstract}


\section{Introduction}
The prosperity of Esport catalyzes a significant research interest for academics~\cite{yu2018fine, xugccegame, tanaka2021lol} and industrial researchers~\cite{johnson2019impacts, whyesports} to pay attention to research topics of Esport’s game commentary since the game commentator could entertain the audiences by interactively informing them of game-related information during live-streaming.
Moreover, the task of collaborative commentary generation aims to generate an expected follow-up commentary based on commentary given by the human commentator to collaborate with him or her interactively.

Although the video caption of Esport games which were collected from the public video website that could be used for game commentary, is processed and punctuated, we notice the punctuated text sequences are usually incomplete game commentary (see Case \textit{Solo} in Table \ref{tab:example}, game scene screenshots were captured from Esports game videos derived from  \textit{Youtube}\footnote{https://www.youtube.com/watch?v=pJRjqqajKwU}\footnote{https://www.youtube.com/watch?v=buNTNOgE2Bs}\footnote{https://www.youtube.com/watch?v=q7NvQUzE0FA}).
Thus, we assume the use of original punctuated sentences will cause the problem of generating incomplete commentary, resulting in the game AI commentator output incomplete game commentary, fails to collaborate with the human commentator. 

To this end, further punctuating sentences is needed to leverage such data effectively. In this paper, we present two strategies for sentence punctuation for game commentary. We compared our two strategies with baseline by employing and fine-tuning a state-of-the-art generative language model Text-to-Text Transfer Transformer (\textit{T5})~\cite{t5} to generate collaborative commentary, respectively. Objective evaluations from automatic metrics and subjective analyses on generated commentaries among three strategies shown our strategy of punctuating sentences by two text sequences outperformed the baseline. 

\begin{table*}[t]
\tiny
    \caption{Examples of punctuating commentary text using 3 sentence punctuation strategies respectively, temporary commentaries derived from \textit{Youtube} given from top to bottom in row Text. \faThumbsOUp\ indicates such sentence is a complete commentary and \faThumbsODown\ vice versa.}
    \label{tab:example}
       \hbox to\hsize{\hfil
  \begin{tabular}{|p{0.5cm} |p{4.3cm} p{0.2cm} |p{4.3cm} p{0.2cm}| p{4.3cm} p{0.2cm}|}
    \hline
Scene
     & 
    \multicolumn{1}{c}{
    \begin{minipage}{4.5cm}
      \includegraphics[width=\linewidth]{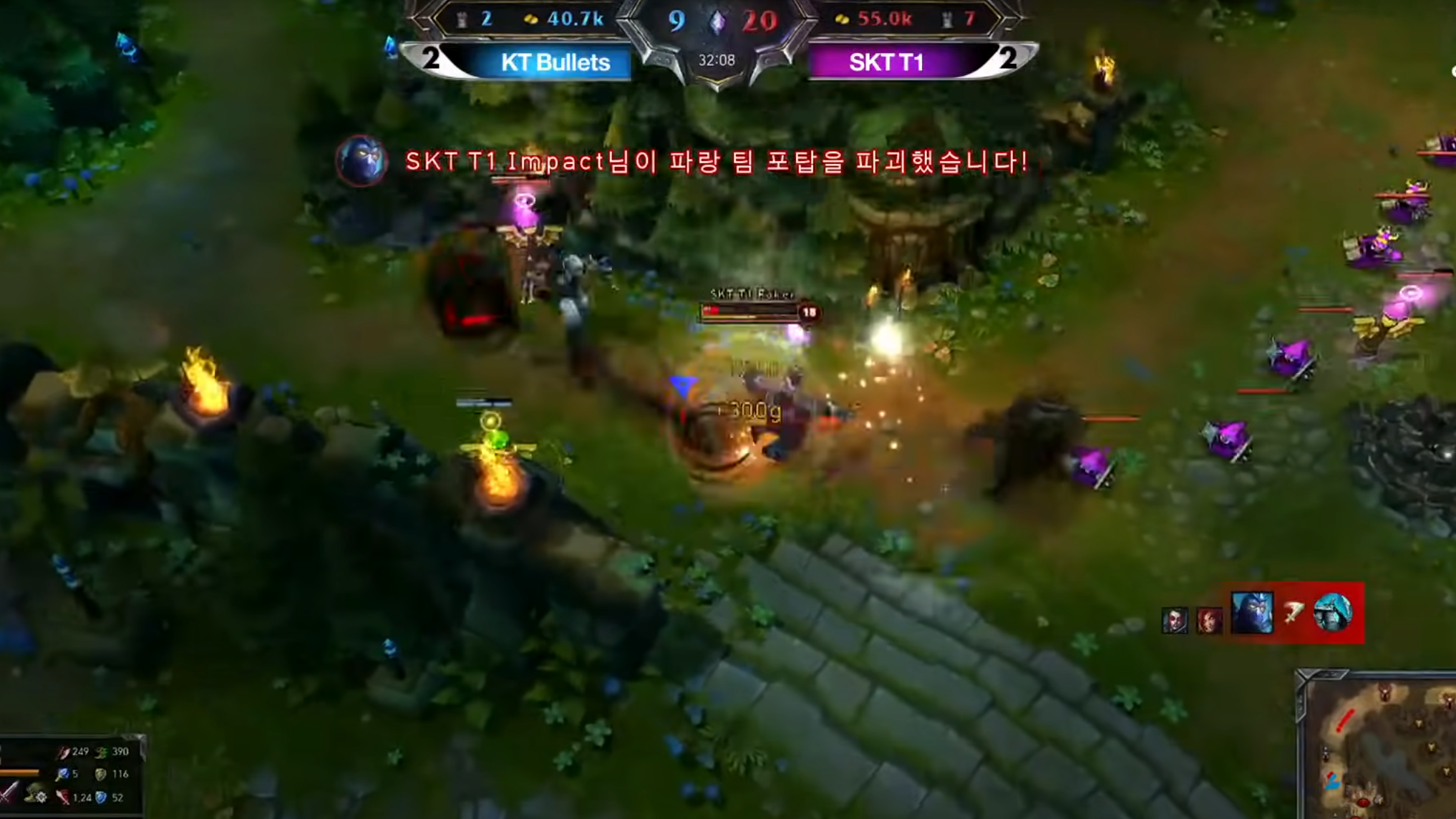}
    \end{minipage} 
    } 
    & &
    \multicolumn{1}{c}{
        \begin{minipage}{4.5cm}
      \includegraphics[width=\linewidth]{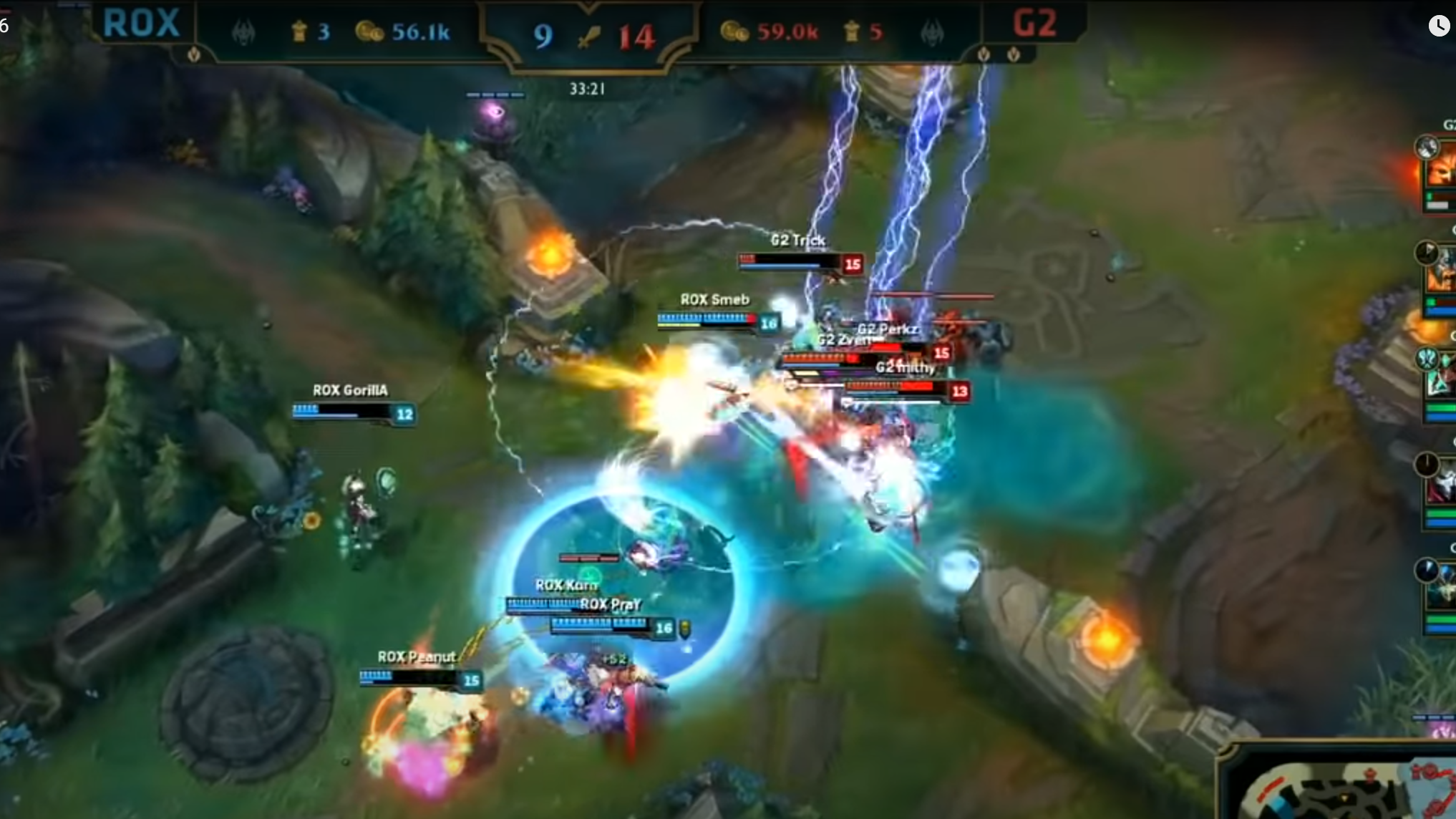}
    \end{minipage}    
    } 
    & &
    \multicolumn{1}{c}{
        \begin{minipage}{4.5cm}
      \includegraphics[width=\linewidth]{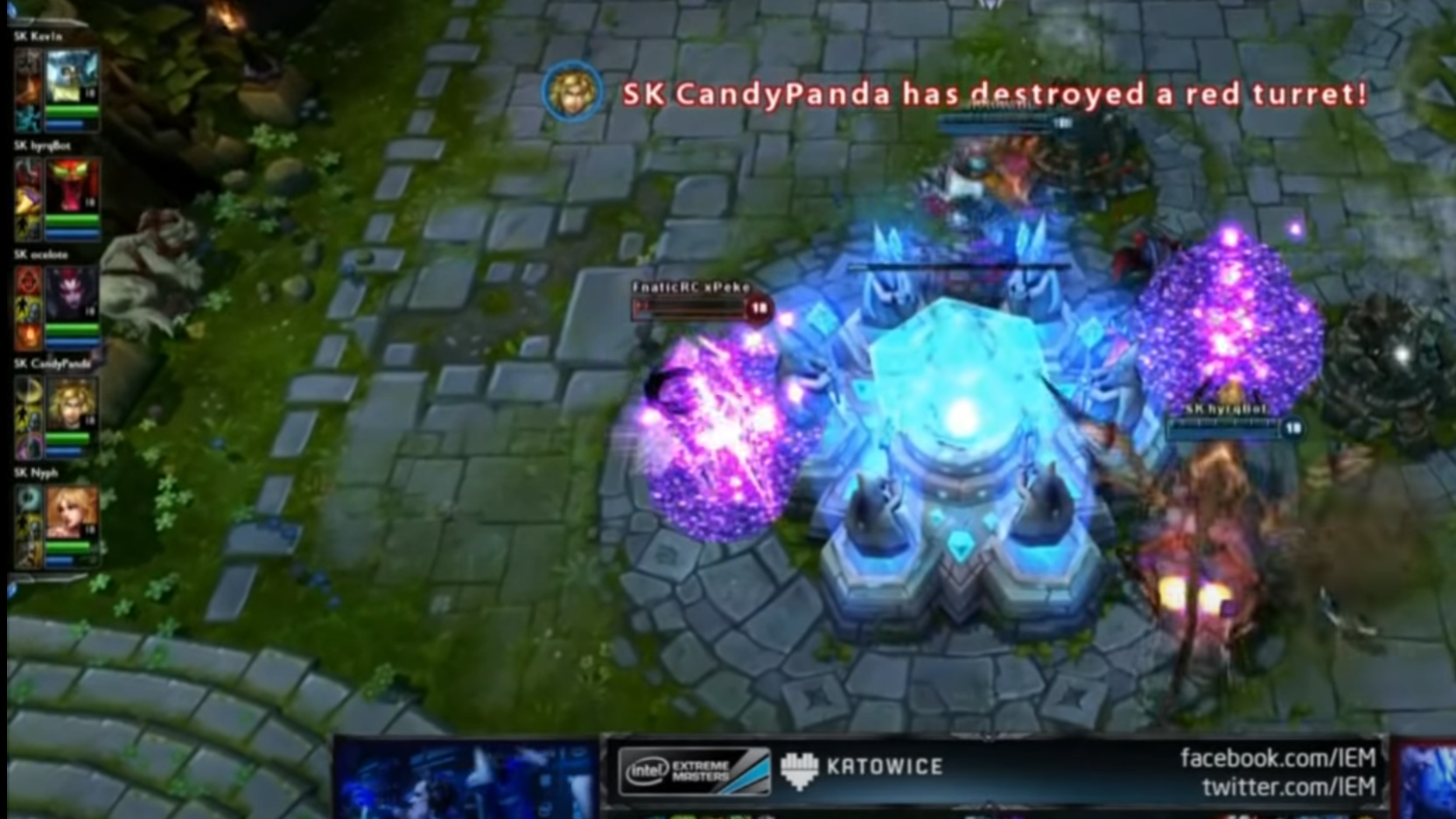}
    \end{minipage}    
    } &
    \\
    \hline
    Case & \multicolumn{2}{c|}{\textbf{Solo}, originally punctuated by \textit{Youtube}} & \multicolumn{2}{c|}{\textbf{Duo}, punctuated by every 2 sentences in \textbf{Solo}} &\multicolumn{2}{c|}{\textbf{Tri}, punctuated by every 3 sentences in \textbf{Solo}} \\
    \hline
Text & \multicolumn{1}{r}{\textless start\textgreater really frightens me but at the same time \textless end\textgreater} & \multicolumn{1}{r|}{\faThumbsODown} & \multicolumn{1}{r}{\textless start\textgreater their strengths of being able to burst} &  & \multicolumn{1}{r}{\textless start\textgreater many differences down and they're just}& \\
 & \multicolumn{1}{r}{\textless start\textgreater I respect that you have just faker \textless end\textgreater} & \multicolumn{1}{r|}{\faThumbsOUp} & \multicolumn{1}{r}{down the bear \textless end\textgreater} & \multicolumn{1}{r|}{\faThumbsOUp} & \multicolumn{1}{r}{gonna file straight up towards those} & \\
 & \multicolumn{1}{r}{\textless start\textgreater nightmares you wake up in a cold sweat \textless end\textgreater} & \multicolumn{1}{r|}{\faThumbsODown} & \multicolumn{1}{r}{\textless start\textgreater and they're hoping to get rocks to pull} &  & \multicolumn{1}{r}{super minions in the base you can see is \textless end\textgreater} & \multicolumn{1}{r|}{\faThumbsOUp}\\
 & \multicolumn{1}{r}{\textless start\textgreater you're like fakers behind me I know \textless end\textgreater} & \multicolumn{1}{r|}{\faThumbsOUp} & \multicolumn{1}{r}{up gets rooted teleports coming in \textless end\textgreater} & \multicolumn{1}{r|}{\faThumbsOUp} & \multicolumn{1}{r}{\textless start\textgreater going in that pack it's definitely}& \\
 & \multicolumn{1}{r}{\textless start\textgreater right like even though I only have a mat \textless end\textgreater} &  \multicolumn{1}{r|}{\faThumbsODown} & \multicolumn{1}{r}{\textless start\textgreater curtain calling wish have already been} &  & \multicolumn{1}{r}{awkward the Nexus Kevin is gonna be out} & \\
  & \multicolumn{1}{r}{\textless start\textgreater on the floor I think is in the bed \textless end\textgreater} & \multicolumn{1}{r|}{\faThumbsODown} & \multicolumn{1}{r}{used a trick is running away \textless end\textgreater} & \multicolumn{1}{r|}{\faThumbsOUp} &\multicolumn{1}{r}{in the bank \textless end\textgreater} & \multicolumn{1}{r|}{\faThumbsODown}\\
 
    \hline
    \end{tabular}\hfil}
\end{table*}

\section{METHOD}

\noindent \textbf{Sentence Punctuation for Collaborative Commentary}: According to the finding mentioned in introduction, the generated commentary should be a complete sentence of commentary rather than a sliced snippet. As shown in Table \ref{tab:result}, in this paper we compared three strategies: the baseline Case \textit{Solo}, Case \textit{Duo} and Case \textit{Tri}, denoting punctuating sentences by 1, 2 and 3 text sequence(s) originally punctuated by \textit{Youtube} respectively.

\noindent \textbf{Pre-trained Generative Language Models:} We used and fine-tuned a state-of-the-art generative pre-trained model, Text-to-Text Transfer Transformer (\textit{T5}) for game collaborative commentary generation.

\section{Experiments}

\subsection{Experimental Settings}\label{AA}
We collected text sequences from 100 videos of one of the most watched Esports games~\cite{newzoo} called \textit{League of Legend} from \textit{Youtube}, further processed into Case \textit{Solo}, \textit{Duo} and \textit{Tri} respectively. Text sequences from 90 videos were used for fine-tuning and others were used for testing. In experiment, we used the small (T5-small) and base (T5-base) version of \textit{T5} model and compared three strategies of sentence punctuation for Esports collaborative commentary generation.

\subsection{Evaluation Results}

\noindent \textbf{Objective Results Given by Automatic Metrics:} We evaluated the generated commentary by using BLEU~\cite{bleu}, ROUGE-1, 2, L~\cite{rouge}, and METEOR~\cite{meteor}. As expected, the performance of ROUGE-1, 2, L among three strategies of sentence punctuation increased, which implies fine-tuning increases the recall rate of words in generated commentary from reference commentary, and in Case \textit{Solo} the METEOR score also increases, which stands for correlation of wording and phrasing with generated text with reference text. We marked the highest score of each metric for three strategies in bold, respectively.

\noindent \textbf{Subjective Analyses on Generated Commentaries:} We compared the generated commentary from fine-tuned T5-base model. We observed (1) the model generated sliced key game-related phrases and incomplete sentences for commentary that is semantically ambiguous in all cases, especially in Case \textit{Solo}. (2) Case \textit{Duo} and Case \textit{Tri} model could generate the entire key game-related phrases such as the game-related scenes narrative with the correct name of characters and moves. (3) However, Case \textit{Tri} model generates more than one commentary (we expect only to generate one commentary for one commentary given by a human commentator).

\begin{table}[t]
\tiny
    \centering
  \caption{Results of conducted comparative experiments. The evaluation was given by automatic metrics on generated collaborative commentary used in this work, respectively.}
  \label{tab:result}
  \begin{tabular}{|l|r|r|r|r|r|}
    \hline

    Model & BLEU & ROUGE-1 & ROUGE-2 & ROUGE-L & METEOR \\
    \hline  
Case \textit{Solo}& \multicolumn{5}{|r|}{}\\
    \hline  
T5-small-solo  &  2.88        & 0.17 &  0.01  &  0.17   &  6.44    \\
T5-base-solo   & 2.88        & 0.52 &  0.07  &  0.52   &  19.17  \\
Fine-tuned-T5-small-solo & 2.92 &  3.15  & \textbf{0.71}    &    3.15 & 7.65 \\
Fine-tuned-T5-base-solo & \textbf{2.97} &  \textbf{3.91}  &  0.48   &  \textbf{3.91}   & \textbf{22.76} \\
    \hline  
    Case \textit{Duo} & \multicolumn{5}{|r|}{} \\
    \hline  
T5-small-duo            & \textbf{1.13} & 0.82    & 0.01    & 0.82    & 12.24  \\
T5-base-duo            & 1.01 & 1.11    & 0.01    & 1.11    & 12.02  \\
Fine-tuned-T5-small-duo & 1.04 & 11.26   & 6.08    & 11.24   & \textbf{13.18}  \\
Fine-tuned-T5-base-duo  & 1.05 & \textbf{16.60}   & \textbf{8.99}    & \textbf{16.52}   & 11.76  \\
    \hline  
    Case \textit{Tri}& \multicolumn{5}{|r|}{}\\
    \hline   
T5-small-tri            & \textbf{1.06} & 1.12    & 0.03    & 1.11    & 12.26  \\
T5-base-tri              & 1.01 & 0.94    & 0.02    & 0.93    & \textbf{12.87}  \\
Fine-tuned-T5-small-tri & 1.03 & 9.32    & 4.61    & 9.30    & 12.84 \\
Fine-tuned-T5-base-tri  & 1.04 & \textbf{11.90}   & \textbf{5.91}    & \textbf{11.89}   & 12.24  \\
    \hline  
\end{tabular}
\end{table}

\section{Conclusion}
To solve the sentence punctuation problem for collaborative commentary generation in Esports live-streaming, this paper presents two strategies for sentence punctuation for punctuating text sequences of game commentary.We conducted comparative experiments among baseline and two strategies on state-of-the-art pre-trained language models for Esports collaborative commentary generation.
From experiment results, we found that (1) the fine-tuning on pre-trained language models improves recalling the words from reference commentary.
(2) Objective evaluations from automatic metrics and subjective analyses on generated examples on generated commentaries among three strategies showed Case \textit{Duo}, the strategy of punctuating sentences by two text sequences outperformed the baseline. (3) From the subjective analyses on Case \textit{Tri} we found that inputting lengthy text sequences results in the AI commentator output more than one commentary that confuses the human commentator.
Our future work will develop a more advanced sentence punctuation strategy study using human evaluation for collaborative commentary generation in Esports live-streaming.

\bibliographystyle{IEEEtran}
\bibliography{main}

\end{document}